\documentclass[10pt,twocolumn,letterpaper]{article}

\usepackage[pagenumbers]{cvpr} % To force page numbers, e.g. for an arXiv version

\usepackage{graphicx}
\usepackage{amsmath}
\usepackage{amssymb}
\usepackage{booktabs}

\usepackage{nicefrac}       % compact symbols for 1/2, etc.
\usepackage{microtype}      % microtypography
\usepackage{array}
\usepackage[ruled, vlined]{algorithm2e}
\usepackage{amsmath}
\usepackage{amssymb}
\usepackage{amsthm}
\usepackage{bm}
\usepackage{multirow}
\usepackage{url}            % simple URL typesetting
\usepackage{xcolor}
\usepackage{xspace}
\usepackage{arydshln}
\usepackage{caption}
\usepackage{subcaption}

\newcommand{\sysname}{DRESS\xspace}
\newcommand\figref[1]{Fig.~\ref{#1}}

\newcommand\secref[1]{Sec.~\ref{#1}}
\newcommand\equref[1]{Eq.(\ref{#1})}

\newcommand{\fakeparagraph}[1]{\vspace{1mm}\noindent\textbf{#1}}

\usepackage[pagebackref,breaklinks,colorlinks]{hyperref}

\usepackage[capitalize]{cleveref}
\crefname{section}{Sec.}{Secs.}
\Crefname{section}{Section}{Sections}
\Crefname{table}{Table}{Tables}
\crefname{table}{Tab.}{Tabs.}

\def\cvprPaperID{15} % *** Enter the CVPR Paper ID here
\def\confName{CVPR}
\def\confYear{2022}

\begin{document}

\title{DRESS: Dynamic REal-time Sparse Subnets}

\author{Zhongnan Qu \thanks{Works are done when he was an intern at Meta.}, Syed Shakib Sarwar, Xin Dong, Yuecheng Li, Ekin Sumbul, Barbara De Salvo \\
Meta Reality Labs Research, US\\
{\tt\small quz@ethz.ch, \{shakib7, yuecheng.li, ekinsumbul, barbarads\}@fb.com, xindong@g.harvard.edu}
}
\maketitle

%%%%%%%%% ABSTRACT
\begin{abstract}
The limited and dynamically varied resources on edge devices motivate us to deploy an optimized deep neural network that can adapt its sub-networks to fit in different resource constraints. However, existing works often build sub-networks through searching different network architectures in a hand-crafted sampling space, which not only can result in a subpar performance but also may cause on-device re-configuration overhead. In this paper, we propose a novel training algorithm, Dynamic REal-time Sparse Subnets (DRESS). DRESS samples multiple sub-networks from the same backbone network through row-based unstructured sparsity, and jointly trains these sub-networks in parallel with weighted loss. DRESS also exploits strategies including parameter reusing and row-based fine-grained sampling for efficient storage consumption and efficient on-device adaptation. Extensive experiments on public vision datasets show that DRESS yields significantly higher accuracy than state-of-the-art sub-networks. 
\end{abstract}

%%%%%%%%% BODY TEXT

\section{Introduction}
\label{sec:introduction}

There is a growing interest to deploy deep neural networks (DNNs) on resource-constrained edge devices to enable new intelligent services such as mobile assistants, augmented reality, etc.
However, state-of-the-art DNNs often make significant demands on memory, computation, and energy. 
Extensive works \cite{bib:ICLR16:Han,bib:ECCV16:Rastegari,bib:NIPS15:Courbariaux,bib:ICLR19:Frankle} have proposed to first compress a pretrained model given resource constraints, and then deploy the compressed model for on-device inference. 

However, the time constraints of many practical embedded systems may dynamically change at run-time, \eg detecting hand position with different speed in real-time, autonomous vehicles' reaction time on city roads and highways. 
On the other hand, the available resources on a single device may also vary, \eg the battery energy, the amount of allocatable RAM.
All these considerations indicate that the deployed inference model should maintain a dynamic capacity to be executed under different resource constraints.

Making DNNs adaptable on resource-constrained edge devices is even more challenging.
Existing works either fail to adapt to different resource constraints, or result in a subpar performance. 
Traditional compression techniques, \eg pruning, quantization, only result in a static inference model. 
Although the compressed model is well-optimized and deployed on edge devices, it can not meet various resource requirements. 
As an alternative, we may deploy for example multiple networks with different sparsity levels \cite{bib:arXiv19:Yu,bib:ICCV19:Liu}, which however need several times more storage consumption in comparison to a single sparse network.
Recent works \cite{bib:ICLR19:Yu,bib:ICLR20:Cai} show that the sub-networks from a pretrained backbone network can reach a decent performance compared to the sub-networks trained individually from scratch.
Nevertheless, they only sample the sub-network architectures along hand-crafted structured dimensionalities, \eg width, kernel size, which leads to sub-optimal results.
Switching among different architectures in run-time may also cause extra re-configuration overhead.

In this paper, we propose Dynamic REal-time Sparse Subnets (DRESS).
\sysname samples sub-networks from the backbone network through row-based unstructured sparsity, while ensuring that nonzero weights of the higher sparsity networks are reused by the lower sparsity networks.
This way, the overall memory consumption is bounded by the network with the lowest sparsity and does not depend on the number of networks, resulting in memory efficiency; all sparse sub-networks leverage the same architecture as the backbone network, leading to re-configuration efficiency.
The sub-network with a higher sparsity (\ie fewer nonzero weights) needs a smaller amount of on-device memory fetching and fewer multiply-accumulate operations (FLOPs), thus shall be adopted to inference under more severe resource constraints, \eg lower energy budget, limited inference time.

Specifically, we (\textit{i}) sample weights w.r.t. their magnitudes in a row-based fine-grained manner; (\textit{ii}) train all sampled sparse sub-networks with weighted loss in parallel; 
(\textit{iii}) further fine-tune batch normalization for each sub-network individually.   
Our contributions are summarized as,
\begin{itemize}
\itemsep -2pt
    \item We design a training pipeline \sysname that generates multiple sub-networks from the backbone network with weight sharing. 
    \item We propose a row-based fine-grained sampling that allows subnets to be efficiently stored and executed using our proposed compressed sparse row (CSR) format.   
    \item Experimental results show \sysname reaches a similar accuracy while only requiring 50\%-60\% disk storage as unstructured pruning.    
\end{itemize}

\section{Related Works}
\label{sec:related}

\fakeparagraph{Network compression \& deployment.}
Network compression focuses on trimming down the DNN model size. 
Commonly used compression techniques consist of, (\textit{i}) designing efficient network architectures manually \cite{bib:arXiv17:Howard,bib:CVPR18:Sandler} or automatically \cite{bib:ICCV19:Howard,bib:ICML19:Tan,bib:ICLR20:Cai}; (\textit{ii}) quantizing weight into lower bitwidth \cite{bib:NIPS15:Courbariaux,bib:ECCV16:Rastegari,bib:CVPR20:Qu}; (\textit{iii}) structured \cite{bib:ICCV19:Liu,bib:ECCV20:Li}/unstructured \cite{bib:ICLR16:Han,bib:ICLR19:Frankle,bib:ICLR20:Renda,bib:ICML21:Evci,bib:NIPS21:Peste} pruning unimportant weights as zeros to reduce the number of operations (also compute energy) and the number of nonzero weights (also memory consumption).
The compressed model is further optimized by some libraries to speed up inference on certain edge platforms, \eg XNNPACK for ArmV7 CPU \cite{bib:XNNPACK}. 
Note that the optimized model often only supports a static computation graph due to the limited resources on edge devices \cite{bib:CMSIS-NN,bib:XNNPACK,bib:Vela}.
We focus on unstructured pruning, since (\textit{i}) it often yields a higher compression ratio \cite{bib:ICLR20:Renda}; (\textit{ii}) the networks with different unstructured sparsity may share the same network architecture, \ie the same computation graph.
Furthermore, the recently released XNNPACK library includes fast kernels for sparse matrix-dense matrix multiplication, which enables sparse DNN acceleration on edge platforms \cite{bib:CVPR20:Elsen}.

\fakeparagraph{Dynamic networks and anytime networks.}
Dynamic networks and anytime networks aim at an efficient inference through adapting network structures. 
Dynamic networks \cite{bib:PAMI21:Han,bib:ICLR18:Huang,bib:CVPR21:Li,bib:CVPR20:Verelst} realize an input-dependent adaptation to reduce the average resource consumption during inference.
Unlike dynamic networks, anytime networks refer to the network whose sub-networks can be executed separately under a resource constraint while achieving a satisfactory performance. 
\sysname falls into the same scope of anytime networks.
MSDNet \cite{bib:ICLR18:Huang} densely connects multiple convolutional layers in both depth direction and scale direction, such that the computation can be saved by early-exiting from a certain layer.
Slimmable networks \cite{bib:ICLR19:Yu,bib:ICCV19:Yu} propose to train a single model which supports multiple width multipliers (\ie number of channels) in each layer.
Subflow \cite{bib:RTAS20:Lee} executes only a sub-graph of the full DNN by activating partial neurons given the varied time constraints.
State-of-the-art anytime networks always sample sub-networks from the backbone network along hand-crafted structured dimensionalities, \eg depth, width, kernel size, neuron.
As zero weights have no effects on the calculation, anytime networks actually perform structured pruning on the backbone network, which could result in a subpar performance in comparison to unstructured sampling.
In addition, resulted sub-networks often have different architectures, \eg different kernel sizes, the re-configuration of the computation graph may bring extra overhead during on-device adaptation.

\section{Dynamic Real-time Sparse Subnets}
\label{sec:DRESS}

\fakeparagraph{Problem Definition}
We aim at sampling multiple subnets from a backbone network.
The backbone network is a traditional DNN consisting of $L$ convolutional (conv) layers or fully connected (fc) layers. 
To achieve memory efficiency, the nonzero weights of the subnet with a higher sparsity are reused by the subnet with a lower sparsity.
This way, we only need to store a table for the lowest sparsity network, including its nonzero weights sorted w.r.t. importance and corresponding indices.
Accordingly, the other networks can be built from the top important weights through a pre-defined sparsity level.
Assume that we sample altogether $K$ sparse subnets, then the preliminary problem is defined as,
\vspace{-0.5em}
\begin{eqnarray}
    \min_{\bm{w},\bm{m}_k}      & \mathcal{L}(\bm{w}\odot\bm{m}_k)      & \forall k \in 1...K       \label{eq:loss} \\
    \text{s.t.}                 & \|\bm{m}_k\|_0 = (1-s_k) \cdot I      & \forall k \in 1...K       \label{eq:constraints1} \\
                                & \bm{m}_i \odot \bm{m}_j = \bm{m}_j    & \forall 1 \le i<j \le K   \label{eq:constraints2} 
\end{eqnarray}
where $\bm{w}$ stands for the weights of the (dense) backbone network; $\bm{m}_k$ stands for the binary mask of the $k$-th subnet; $s_k$ stands for the pre-defined sparsity level. 
$\mathcal{L}(.)$ denotes the loss function, $\|.\|_0$ denotes the L0-norm, $\odot$ denotes the element-wise multiplication. 
Note that $\bm{w}\in\mathbb{R}^I$, $\bm{m}_k\in\{0,1\}^I$, where $I$ is the total number of weights. 
We have $0<s_1<...<s_K\le1$. 
$\bm{w}_k$ is denoted as nonzero weights of the $k$-th sparse subnet, \ie $\bm{w}_k=\bm{w}\odot\bm{m}_k$.

\begin{figure*}[htbp!]
     \centering
     \begin{subfigure}[b]{0.6\linewidth}
         \centering
         \includegraphics[width=\textwidth]{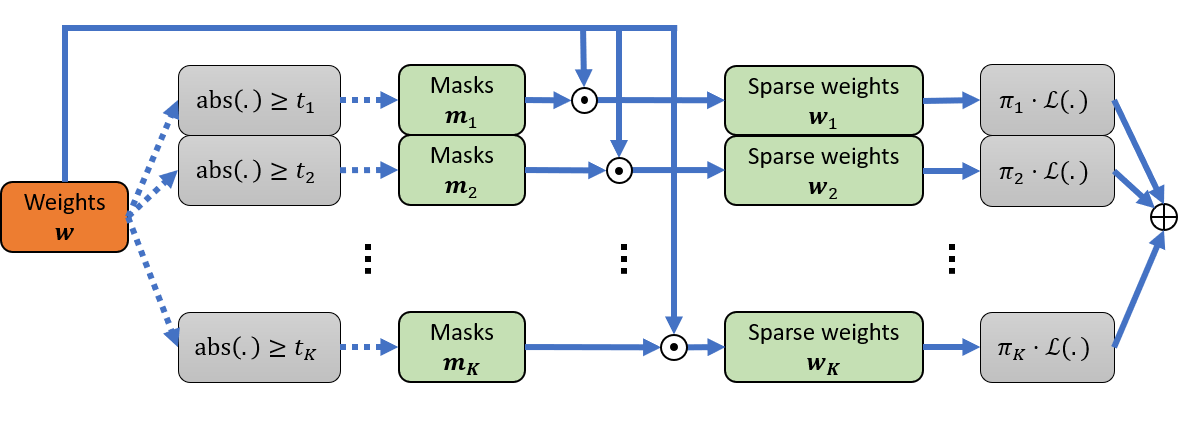}
     \end{subfigure}
     \hfill
     \begin{subfigure}[b]{0.39\linewidth}
         \centering
         \includegraphics[width=\textwidth,height=0.60\linewidth]{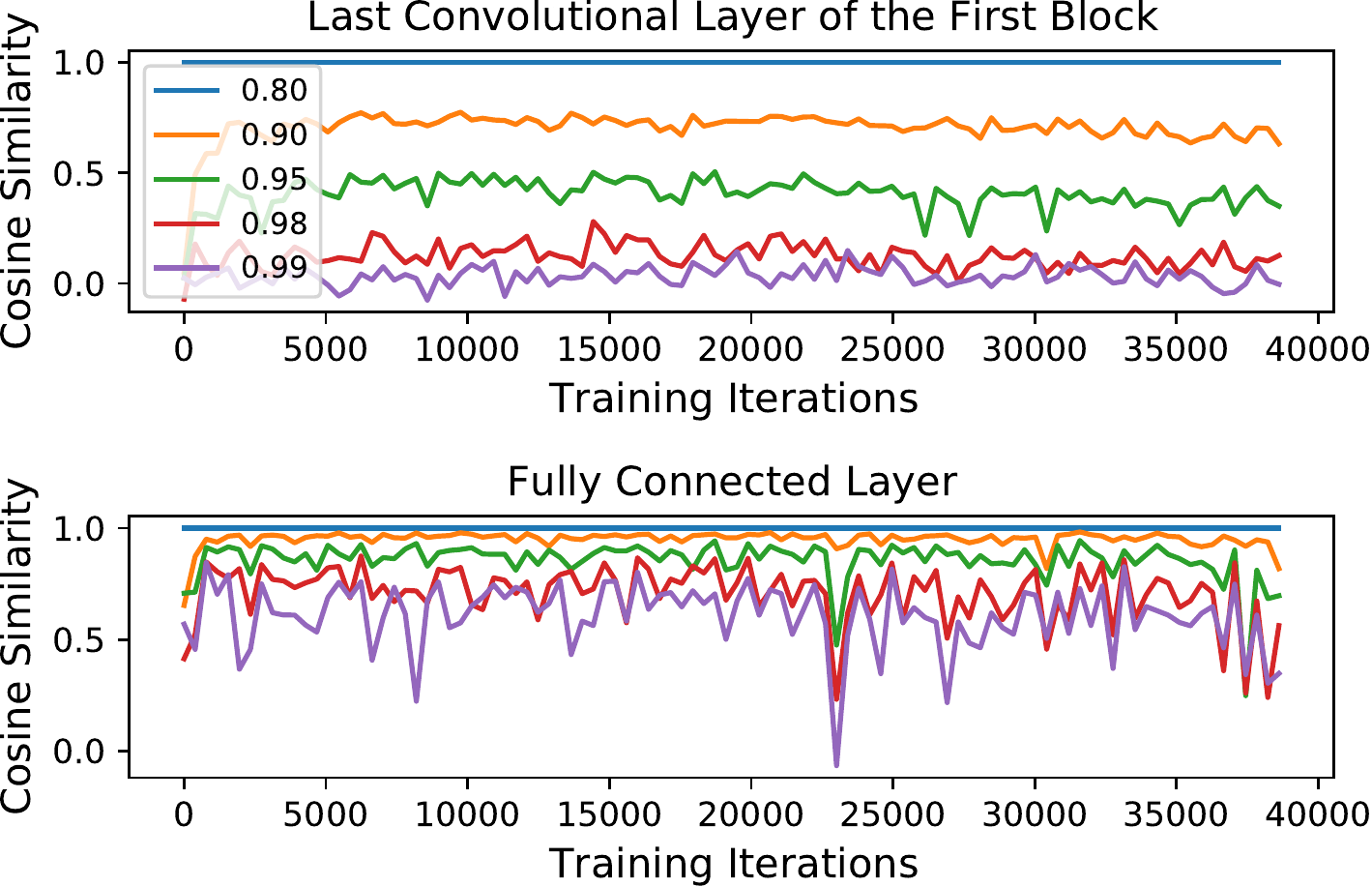}
     \end{subfigure}
     \vspace{-0.5em}
    \caption{Left: The computation graph used in parallel training multiple subnets. 
    The orange blocks are the leaf variables to be optimized; the green blocks are the intermediate variables; the gray blocks are the computation units. 
    The solid arrows are the differentiable operation to be backpropagated, but the dashed arrows are not.
    Right: The cosine similarity between the loss gradients of 5 subnets (with sparsity 0.8,0.9,0.95,0.98,0.99) and that of the lowest sparsity subnet with sparsity 0.8.
    We show two typical layers in ResNet20.}
    \label{fig:process}
    \vspace{-0.7em}
\end{figure*}

In the following sections, we detail how to solve \equref{eq:loss}-\equref{eq:constraints2} in our \sysname algorithm. 
\sysname consists of three training stages, (\textit{i}) dense pre-training, where the backbone network is trained from scratch to provide a good initial point for the following sparse training; 
(\textit{ii}) \sysname training, where multiple sparse subnets are sampled from the backbone network (\secref{sec:sample}, \secref{sec:store}) and are jointly trained in parallel with weighted loss (\secref{sec:optimize}); 
(\textit{iii}) post-training on batch normalization (BN), where BN layers are further optimized individually for each subnet to better reveal the statistical information, as BN layers often require a rather small amount of memory and computation.  
The overall pseudocode is shown in Alg.~\ref{alg:DRESS} in Appendix~\ref{app:pseudocodes}. 

\subsection{How to Sample Sparse Subnets}
\label{sec:sample}

Unlike traditional anytime networks that sample subnets along structured dimensionalities, \sysname samples subnets weight-wise which extremely enlarges the sampling space.
The naive approach could be iteratively sampling $K$ subnets by exhaustively searching the best-performed subnet inside the current subnet. 
However, it can be either conducted rarely or infeasible due to the high complexity.
To reduce the complexity, we propose to greedily sample the subnets based on the importance of weights.
As previous pruning works \cite{bib:ICLR16:Han,bib:ICLR19:Frankle,bib:ICLR20:Renda}, the importance of weights is measured by their magnitudes. 
Given an overall sparsity level $s_k$, the $(1-s_k)\cdot I$ weights with the largest magnitudes will be sampled to build the subnet.
To further reduce the sorting complexity, the weights are only sorted inside each layer according to a layer-wise sparsity $s_{k,l}$, where $l$ denotes the layer index.
The global sorting, also the (re-)allocation of layer-wise sparsity, is conducted only if the average accuracy of subnets does not improve anymore, see Alg.~\ref{alg:DRESS}.
During (re-)allocation, the weights of all layers with the largest magnitudes are selected in sequence until reaching the overall sparsity $s_k$, and the layer-wise sparsity can be then calculated accordingly. 

\subsection{How to Optimize Subnets}
\label{sec:optimize}

With sampled binary masks, we can now build and train the subnets. 
Our concept for optimizing subnets is based on the key insight: \textit{in comparison to iterative training of subnets in progressively decreased/increased sparsity, parallel training allows multiple subnets to be sampled and optimized jointly, which avoids being stuck into bad local optimum and thus yields higher performance.}
Experimental results in Appendix \ref{app:iter_para} verify that parallel training yield significantly higher performance than iterative training. 
In parallel training, \equref{eq:loss} can be re-written as,  
\vspace{-0.5em}
\begin{equation}
    \min_{\bm{w},\bm{m}_k}~~~\sum_{k=1}^K \pi_k \cdot \mathcal{L}(\bm{w}\odot\bm{m}_k) 
    \label{eq:loss_parallel}
    \vspace{-0.5em}
\end{equation}
where $\pi_k$ is the normalized scale ($\sum_{k=1}^K \pi_k = 1$) used to weight $K$ loss items, which will be discussed later. 
This process determines a threshold $t_k$, the mask value $m_{k,i}=1$ if $\mathrm{abs(}w_i\mathrm{)}\ge t_k$, otherwise 0, $\forall i\in1...I$.
$t_k$ is set to the value such that $(1-s_k)$ of weights have a larger absolute value than $t_k$.
Clearly, we have $t_1<t_2<...<t_K$ due to the constraints of \equref{eq:constraints1}-\equref{eq:constraints2}.
In each training iteration, we sample $K$ sparse subnets $\bm{w}_{1:K}$ and optimize the backbone $\bm{w}$ with the weighted sum of their losses, see in \figref{fig:process} Left.

We parallelly train 5 subnets of ResNet20 \cite{bib:CVPR16:He} on CIFAR10, and let 5 loss items weighted equally, \ie $\pi_{1:5}=0.2$. 
We plot the cosine similarity between the loss gradients (\ie $(\partial\mathcal{L}(\bm{w}_k)/\partial\bm{w}_k) \odot \bm{m}_k$) of 5 subnets and that of the lowest sparsity subnet along with the training iterations, see in \figref{fig:process} Right.
It shows that the loss gradients of different subnets are always positively correlated, which also verified that multiple subnets are jointly trained towards the optimal point in the loss landscape.
Because of \equref{eq:constraints2}, the nonzero weights in higher sparsity subnets are also selected by other subnets, which means these weights are optimized with a larger step size than others.  
To balance the step size, we propose to weight the loss items by the ratio of trainable weights (\ie $1-s_k$) together with a correction factor $\gamma$.
Particularly, $\alpha_k = (1-s_k)^\gamma$, and with the normalization, $\pi_k = \alpha_k/\sum_{k=1}^K\alpha_k$.
Note that $\gamma=0$ means weighting loss items equally. 
Experimentally, we find that $\gamma=0.5$ often yields a satisfactory performance, see Appendix \ref{app:gamma}.

\subsection{How to Store Subnets}
\label{sec:store}

\begin{figure}[tbp!]
	\centering
	\includegraphics[width=0.483\textwidth]{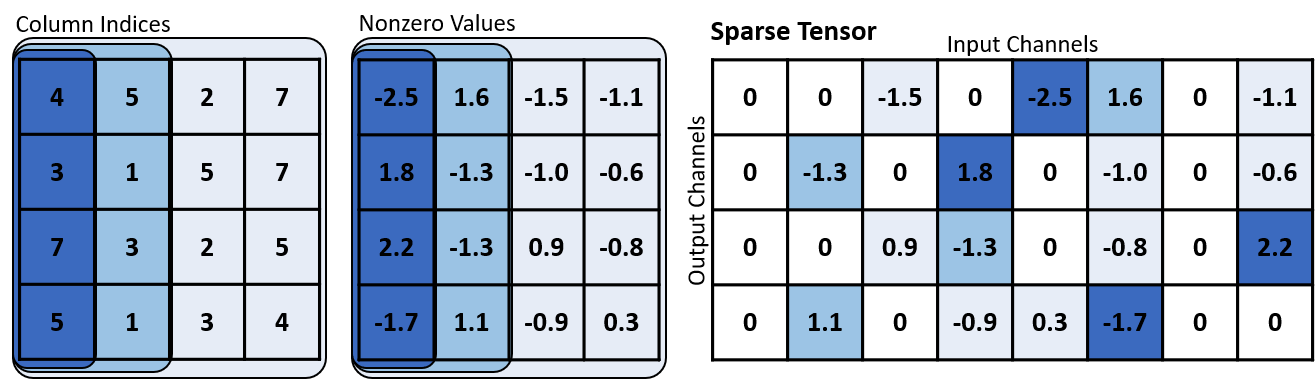}
	\caption{\sysname CSR format of row-based unstructured sparse tensor. 
	The example weight tensor is from a $1\times1$ conv layer with 8 input channels and 4 output channels. 
	Each row has 8 weights in total, also the row size $N=8$.
	There are 3 sparse subnets with sparsity 0.5, 0.75, and 0.875. 
	Each subnet has 4, 2, and 1 nonzero weights per row, respectively.}
	\label{fig:csr2}
	\vspace{-0.6em}
\end{figure}

State-of-the-art libraries often encodes sparse tensor in compressed sparse row (CSR) format for sparse inference \cite{bib:XNNPACK,bib:CVPR20:Elsen}.
To achieve an efficient inference on different sparse subnets while without extra memory overhead, we adopt a \textit{row-based unstructured} sparsity, where different rows leverage the same sparsity level.
Especially, for sparsity $s_k$, all rows have exactly $(1-s_k)\cdot N$ nonzero weights, where $N$ is the number of weights per row.
In comparison to conventional unstructured sparsity, this kind of sparsity (a.k.a. \textit{N:M fine-grained structure sparsity} \cite{bib:ICLR21:Zhou,bib:NIPS21:Hubara,bib:NIPS21:Sun}) can also be accelerated with sparse tensor cores of A100 GPUs \cite{bib:arXiv21:Mishra} for both training and inference, and thus becomes prevailing recently. 
The column indices of nonzero weights are stored in descending order of the importance (also the weight magnitude) in a two-dimensional table.
The nonzero weights are stored in a table with the same order as the column indices.
This \sysname CSR format needs to store (\textit{i}) the subnet with the lowest sparsity including the table of the column indices and the table of nonzero weights, (\textit{ii}) $K$ integers $\{(1-s_k)\cdot N\}_{k=1}^K$. 
When adopting the $k$-th subnet, we fetch the first $(1-s_k)\cdot N$ columns from both tables as shown in \figref{fig:csr2}.
Note that all fetched subnets follow the CSR format ($(1-s_k)\cdot N$ is used to build the row indices in CSR) under the same architecture, which allows us to leverage available libraries to achieve a fast on-device inference without re-configuration overhead.
To obtain \sysname CSR format, the sampling process needs to be adjusted accordingly.  
Especially, for layer $l$, we first pre-define a row size $N_l$ and reshape the weight tensor into rows. 
In this paper, the weights corresponding to each output channel (\eg each conv filter) are formed into one row.  
The row-based sampling is shown in Alg.~\ref{alg:rowsampling} in Appendix~\ref{app:rowsampling}.

\section{Experiments}
\label{sec:experiments}

We implement \sysname with Pytorch \cite{bib:NIPSWorkshop17:Paszke}, and evaluate its performance on CIFAR10 \cite{bib:cifar} using ResNet20 \cite{bib:CVPR16:He}, on ImageNet \cite{bib:ILSVRC15} using ResNet50 \cite{bib:CVPR16:He}. 
We randomly select 20\% of each original test dataset (original validation dataset for ImageNet) as the validation dataset, and the remainder as the test dataset.
We use the Nesterov SGD optimizer with the cosine schedule for learning rate decay in all methods. 
We report the Top-1 test accuracy for the subnets of the epoch when the validation dataset achieves the highest average accuracy over all subnets.
More implementation details are provided in Appendix~\ref{app:implementation}. 
Row-based unstructured sampling is conducted in all layers except for BN layers. 
We set the sparsity levels $s_{1:5}=0.8,0.9,0.95,0.98,0.99$ for ResNet20, and $s_{1:4}=0.5,0.8,0.9,0.95$ for ResNet50.

\begin{figure}[htbp!]
	\centering
	\includegraphics[width=0.48\textwidth,height=0.33\textwidth]{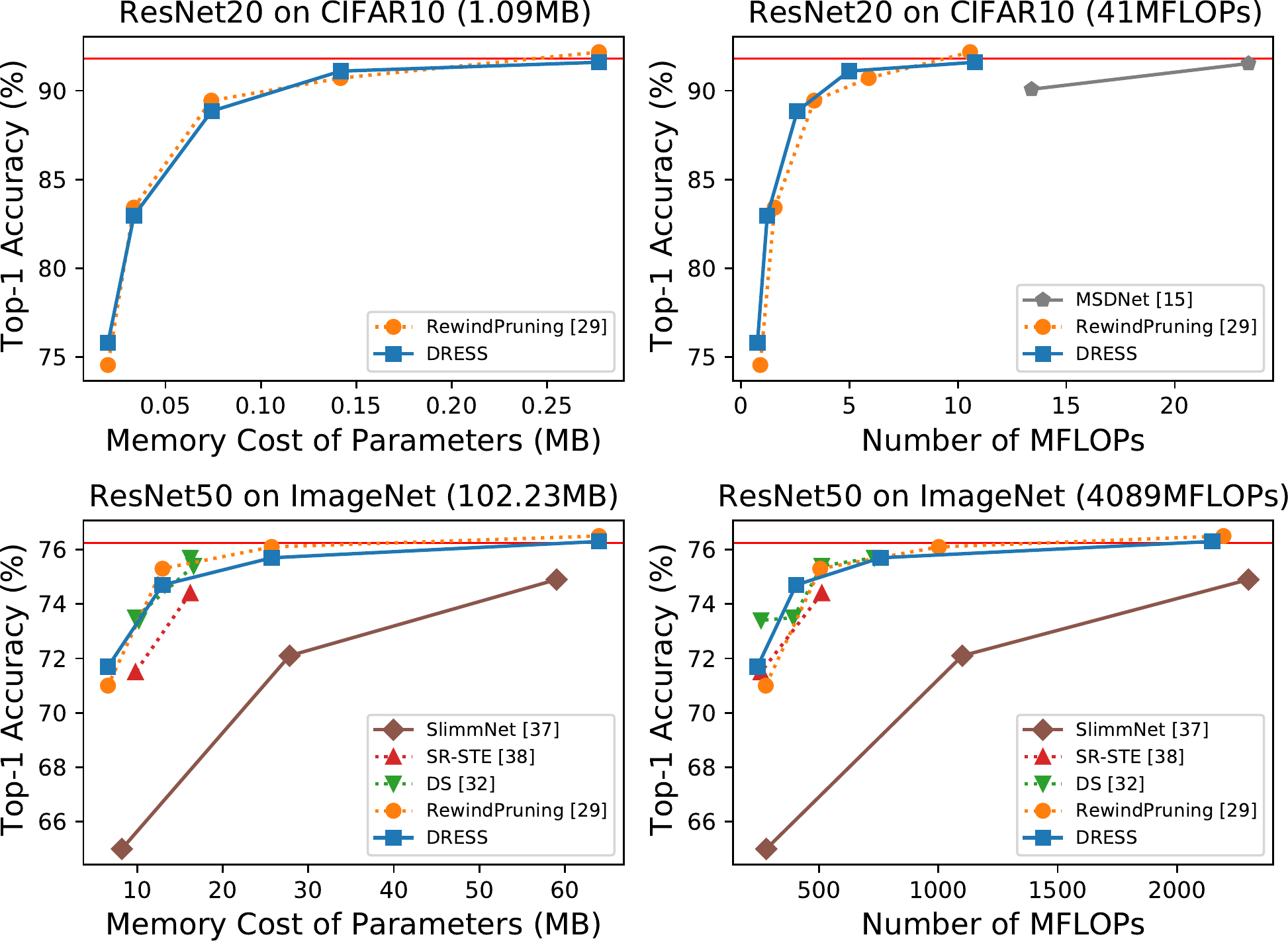}
	\caption{Comparing \sysname with other baselines. The methods that do not involve weight sharing among different networks are plotted with dotted curves. The memory cost and the number of MFLOPs of the original backbone networks are reported in the parentheses in titles; their accuracy is shown as red horizontal lines.}
	\label{fig:benchmark}
	\vspace{-0.7em}
\end{figure}

We compare the performance of the subnets generated by \sysname with various methods, including (\textit{i}) anytime networks \cite{bib:ICLR18:Huang,bib:ICLR19:Yu}, where the sub-networks with different width or depth can be cropped from the backbone network; (\textit{ii}) unstructured pruning \cite{bib:ICLR20:Renda}, where the backbone network is pruned with the same sparsity as \sysname; (\textit{iii}) N:M fine-grained structure pruning \cite{bib:ICLR21:Zhou,bib:NIPS21:Sun}.
We choose two metrics for comparison, the memory cost of parameters and the number of FLOPs.
FLOPs dominate in the entire computation burden, thus fewer FLOPs can (but does not necessarily) result in a smaller computation time.   
The memory cost of parameters represents not only the disk storage consumption but also the amount of memory fetching during on-device inference.
Note that memory access often consumes more time and energy than computation \cite{bib:ISSCC14:Horowitz}.
Assume that each weight uses 32-bit floating point.
\sysname, (\textit{ii}), and (\textit{iii}) generate sparse tensors, thus their memory cost also includes the indices of nonzero weights.
Each index of nonzero weights are encoded into 8-bit in \sysname and (\textit{ii}) as \cite{bib:tensorflow,bib:XNNPACK}, wheres the binary mask is stored for indexing in \cite{bib:ICLR21:Zhou,bib:NIPS21:Sun}. 

The results are plotted in \figref{fig:benchmark}.
\sysname require a significantly lower memory cost and fewer FLOPs than other anytime networks.
Furthermore, different subnets of \sysname leverage the same architecture as the backbone network, which avoids the re-configuration overhead of switching different architectures in other anytime networks \cite{bib:ICLR18:Huang,bib:ICLR19:Yu,bib:ICCV19:Yu}.
Thanks to the weight reusing, the disk storage is only determined by the largest network for both \sysname and anytime networks \cite{bib:ICLR18:Huang,bib:ICLR19:Yu,bib:ICCV19:Yu}.
In comparison to (\textit{ii}), \ie without weight reusing, \sysname reaches a similar accuracy while only requiring 50\%-60\% of disk storage. 

\section{Conclusion}
\label{sec:conclusion}

In this paper, we introduce \sysname. 
\sysname is able to build multiple subnets via row-based unstructured sparsity with weight sharing and architecture sharing. 
The resulted multiple subnets achieve a similar accuracy as state-of-the-art single compressed network, while enabling memory efficiency and configuration efficiency.

%%%%%%%%% REFERENCES
\clearpage
{\small
\bibliographystyle{ieee_fullname}
\bibliography{egbib}
}

%%%%%%%%% APPENDICES
\clearpage
\section*{Appendix}

\appendix

\section{Pseudocodes}
\label{app:pseudocodes}

\subsection{Parallel Training Subnets (\sysname)}
\label{app:parallel}

\sysname consists of three training stages, (\textit{i}) dense pre-training, where the backbone network is trained from scratch with a traditional optimizer to provide a good initial point for the following sparse training; 
(\textit{ii}) \sysname training, where the multiple sparse subnets are sampled from the backbone network and are jointly trained in parallel with weighted loss; 
(\textit{iii}) post-training on batch normalization (BN), where BN layers are further optimized individually for each subnet to better reveal the statistical information. 
The overall pseudocode is shown in Alg.~\ref{alg:DRESS}. 

\begin{algorithm}[htbp!]
\caption{Dynamic REal-time Sparse Subnets}\label{alg:DRESS}
\KwIn{Initial random weights $\bm{w}$, training dataset $\mathcal{D}_\mathrm{tr}$, validation dataset $\mathcal{D}_\mathrm{val}$, overall sparsity $\{s_k\}_{k=1}^K$, normalized loss weights $\{\pi_k\}_{k=1}^K$} 
\KwOut{Optimized weights $\bm{w}$, binary masks $\{\bm{m}_k\}_{k=1}^K$}
\tcc{Dense pre-training}
Train dense network $\bm{w}$ with traditional optimizer\;
\tcc{DRESS training}
Allocate layer-wise sparsity $\{s_{k,l}\}_{l=1}^L$ for each $s_k$\;
Initiate $\bm{w}^0=\bm{w}$\;
\For {$q \leftarrow 1$ \KwTo $Q$} { 
    \tcp{The $q$-th training iteration}
    Fetch mini-batch from $\mathcal{D}_\mathrm{tr}$\;
    Initialize backbone-net gradient $\bm{g}(\bm{w}^{q-1})=\bm{0}$\;
    \For {$k \leftarrow 1$ \KwTo $K$} {
        Sample a subnet with layer-wise sparsity $\{s_{k,l}\}_{l=1}^L$ and get its mask $\bm{m}_k$\;
        Get sparse subnet $\bm{w}^{q-1}_k=\bm{w}^{q-1}\odot\bm{m}_k$\;
        Back-propagate subnet gradient $\bm{g}(\bm{w}^{q-1}_k)=\pi_k\cdot\frac{\partial\mathcal{L}(\bm{w}^{q-1}_k)}{\partial\bm{w}^{q-1}_k}$\;
        Accumulate backbone-net gradient $\bm{g}(\bm{w}^{q-1})=\bm{g}(\bm{w}^{q-1})+\bm{g}(\bm{w}^{q-1}_k)\odot\bm{m}_k$\;
    }
    Compute optimization step $\Delta\bm{w}^{q}$ with $\bm{g}(\bm{w}^{q-1})$\;
    Update $\bm{w}^{q} = \bm{w}^{q-1} + \Delta\bm{w}^{q}$\;
    \uIf{Higher average (epoch) accuracy on $\mathcal{D}_\mathrm{val}$}
    {Save $\bm{w}=\bm{w}^q$ and $\{\bm{m}_k\}_{k=1}^K$\;}
    \Else{Re-allocate layer-wise sparsity $\{s_{k,l}\}_{l=1}^L$ for each $s_k$\;}
}
\tcc{Post-training on batch normalization (BN)}
\For {$k \leftarrow 1$ \KwTo $K$} {
    Load $\bm{w}$ and $\bm{m}_k$\;
    Fine-tune BN layers of subnet $\bm{w}\odot\bm{m}_k$\;
}
\end{algorithm}

\subsection{Iterative Training Subnets}
\label{app:iterative}

Recall that we assume there are $K$ sparsity levels, $s_{1:K}$, where $0<s_1<s_2<...<s_K\le1$.
In iterative training, each subnet is optimized separately in each iteration, also altogether $K$ iterations. 
In each iteration, we mainly adopt the idea of traditional unstructured magnitude pruning \cite{bib:ICLR20:Renda}, which is the current best-performed pruning method aiming at the trade-off between the model accuracy and the number of zero’s weights.
Traditional unstructured pruning \cite{bib:ICLR20:Renda} conducts iterative pruning with a pruning scheduler $p^{1:R}$. 
The network progressively reaches the desired sparsity $s$ until the $R$-th pruning iteration.
We choose $p^{1:5}=0.5,0.8,0.9,0.95,1$, \ie the sparsity is set to $0.5s,0.8s,0.9s,0.95s,s$ in 5 pruning iterations, respectively.
During each pruning iteration, the network is pruned with the corresponding sparsity, and the nonzero weights are sparsely fine-tuned for several epochs with learning rate rewinding \cite{bib:ICLR20:Renda}.

\begin{algorithm}[tbp!]
\caption{Iterative training with increased sparsity}\label{alg:increased}
\KwIn{Initial random weights $\bm{w}$, training dataset $\mathcal{D}_\mathrm{tr}$, validation dataset $\mathcal{D}_\mathrm{val}$, sparsity $\{s_k\}_{k=1}^K$, pruning scheduler $\{p^r\}_{r=1}^{R}$} 
\KwOut{Optimized weights $\bm{w}$, binary masks $\{\bm{m}_k\}_{k=1}^K$}
\tcc{Dense pre-training}
Train dense network $\bm{w}$ with traditional optimizer\;
\tcc{Traditional pruning, also k=1}
\For {$r \leftarrow 1$ \KwTo $R$} { 
    \tcp{The $r$-th pruning iteration}
    Prune with sparsity $s_1\cdot p^r$ and get mask $\bm{m}_1^r$\;
    Sparsely fine-tune nonzero weights $\bm{w}\odot\bm{m}_1^r$ with several epochs on $\mathcal{D}_\mathrm{tr}$\;
}
Get mask $\bm{m}_1 = \bm{m}_1^R$\;
\tcc{Iterative (training)}
\For {$k \leftarrow 2$ \KwTo $K$} { 
    Get the previous subnet $\bm{w}_{k-1} = \bm{w}\odot\bm{m}_{k-1}$\; 
    Sample a subnet from $\bm{w}_{k-1}$ with sparsity $s_k$ and get mask $\bm{m}_k$\;
    \tcp{Note no training here.}
}
\end{algorithm}

\begin{algorithm}[tbp!]
\caption{Iterative training with decreased sparsity}\label{alg:decreased}
\KwIn{Initial random weights $\bm{w}$, training dataset $\mathcal{D}_\mathrm{tr}$, validation dataset $\mathcal{D}_\mathrm{val}$, sparsity $\{s_k\}_{k=1}^K$, pruning scheduler $\{p^r\}_{r=1}^{R}$}
\KwOut{Optimized weights $\bm{w}$, binary masks $\{\bm{m}_k\}_{k=1}^K$}
\tcc{Dense pre-training}
Train dense network $\bm{w}$ with traditional optimizer\;
\tcc{Iterative training}
Set $s_{K+1}=1$ and $\bm{m}_{K+1}=\bm{0}$\;
\For {$k \leftarrow K$ \KwTo $1$} { 
    Get the complementary subnet $\bm{w}^\mathrm{cs}=\bm{w}\odot(1-\bm{m}_{k+1})$\;
    \For {$r \leftarrow 1$ \KwTo $R$} { 
        \tcp{The $r$-th pruning iteration}
        Sample a subnet from $\bm{w}^\mathrm{cs}$ with sparsity $(1-(s_{k+1}-s_k))\cdot p^r$ and get mask $\bm{m}_k^{\mathrm{cs},r}$\;
        Merge mask $\bm{m}_k^{r}=\bm{m}_{k+1}+\bm{m}_k^{\mathrm{cs},r}$\;
        Initiate $\bm{w}^0=\bm{w}$\;
        \For {$q \leftarrow 1$ \KwTo $Q$} { 
            \tcp{The $q$-th training iteration}
            Fetch mini-batch from $\mathcal{D}_\mathrm{tr}$\;
            Get sparse subnet $\bm{w}^{r,q-1}_k=\bm{w}^{q-1}\odot\bm{m}_k^{r}$\;
            Back-propagate subnet gradient $\bm{g}(\bm{w}^{r,q-1}_k)=\frac{\partial\mathcal{L}(\bm{w}^{r,q-1}_k)}{\partial\bm{w}^{r,q-1}_k}$\;
            Compute optimization step $\Delta\bm{w}^{q}$ with $\bm{g}(\bm{w}^{r,q-1}_k)\odot\bm{m}_k^{\mathrm{cs},r}$\;
            Update $\bm{w}^{q} = \bm{w}^{q-1} + \Delta\bm{w}^{q}$\;
        }
        Save $\bm{w}=\bm{w}^Q$
    }
    Save mask $\bm{m}_k=\bm{m}_k^R$\;
}
\end{algorithm}

The pseudocode of training multiple subnets iteratively with increased sparsity is shown in Alg.~\ref{alg:increased}.
With progressively increased sparsity (from $s_1$ to $s_K$), the first optimized subnet of $\bm{w}\odot\bm{m}_1$ already contains all subsequent subnets with higher sparsity due to the constraint of \equref{eq:constraints2} (see in the main text).
The first sparse subnet $\bm{w}\odot\bm{m}_1$ is trained by traditional unstructured pruning \cite{bib:ICLR20:Renda} as discussed above.
In the following iteration $k$ ($k\in2...K$), the subnet with sparsity $s_k$ is directly sampled from the previous subnet without any retraining regarding the constraint of \equref{eq:constraints2} (see in the main text).

The pseudocode of training multiple subnets iteratively with decreased sparsity is shown in Alg.~\ref{alg:decreased}.
For progressively decreased sparsity (from $s_K$ to $s_1$), the sampling and training process only happen in the complementary part of the previous subnet, due to the constraint of \equref{eq:constraints2}. 
Particularly, in iteration $k$ ($k\in K...1$), we should (\textit{i}) sample the new subnet from the backbone network with sparsity $s_k$ that contains the subnet of $\bm{w}\odot\bm{m}_{k+1}$; (\textit{ii}) freeze the subnet of $\bm{w}\odot\bm{m}_{k+1}$ and only update the other weights.
We still adopt the iterative pruning idea when training each subnet as mentioned before, \ie the sparsity of the $k$-th subnet gradually approaches the target sparsity $s_k$.
Note that the dense backbone network is maintained and updated during iterative training subnets, 

\begin{algorithm}[tbp!]
\caption{Row-based unstructured sampling}\label{alg:rowsampling}
\KwIn{Weight tensor $\bm{w}\in\mathbb{R}^{H \times N}$, row size $N$, sparsity $\{s_k\}_{k=1}^K$}
\KwOut{Binary masks $\{\bm{m}_k\}_{k=1}^K$}
\For {$k \leftarrow 1$ \KwTo $K$} {
    Initiate binary mask $\bm{m}_k=0^{H \times N}$\;
    Get the number of nonzero weights per row, $N_k^\mathrm{nz}=N\cdot (1-s_k)$\;
}
\For {$h \leftarrow 1$ \KwTo $H$} { 
    Sort the weight magnitudes of row $\bm{w}_{h,:}$ in descending order\;
    \For {$k \leftarrow 1$ \KwTo $K$} {
        Set the mask value of $\bm{m}_{k,h,:}$ as 1 for Top-$N_k^\mathrm{nz}$ indices\;
    }
}
\end{algorithm}

\subsection{Row-Based Unstructured Sampling}
\label{app:rowsampling}

In this section, we present our algorithm of row-based unstructured sampling, see Alg.~\ref{alg:rowsampling}.
We focus on sampling a weight tensor $\bm{w}$ from a conv layer or a fc layer.
We first define a row size $N$ for the weight tensor.
Note that $N$ must be divisible by the total number of weights in $\bm{w}$.
The weight tensor $\bm{w}$ is then reshaped into the form of $\mathbb{R}^{H \times N}$, \ie $N$ weights per row and $H$ rows in total.
Given $K$ sparsity levels $s_{1:K}$, $K$ binary masks with the form of $\{0,1\}^{H \times N}$ are generated.
Binary masks can be reshaped into the original form of the weight tensor accordingly.

\section{Implementation Details}
\label{app:implementation}

\subsection{ResNet20 on CIFAR10}
\label{app:cifar}

CIFAR10 \cite{bib:cifar} is an image classification dataset, which consists of $32\times32$ color images in 10 object classes. 
Each class contains 6000 data samples.
It contains a training dataset with 50000 data samples, and a test dataset with 10000 data samples. 
We use the original training dataset for training, and randomly select 2000 samples in the original test dataset for validation, and the rest 8000 samples for testing.
We use a mini-batch with a size of 128 training on 1 NVIDIA V100 GPU.

\fakeparagraph{ResNet20.}
The network architecture is the same as ResNet-20 in the original paper~\cite{bib:CVPR16:He}.
ResNet20 needs around 1.09MB (1.11MB for CIFAR100) to store the weights and around 41MFLOPs for a single image inference.
We use the Nesterov SGD optimizer with the cosine schedule for learning rate decay. 
The initial learning rate is set as 0.1; the momentum is set as 0.9; the weight decay is set as 0.0005; the number of training epochs is set as 100.

\subsection{ResNet50 on ImageNet}
\label{app:imagenet}

ImageNet \cite{bib:ILSVRC15} is an image classification dataset, which consists of high-resolution color images in 1000 object classes. 
It contains a training dataset with 1.2 million data samples, and a validation dataset with 50000 data samples. 
Following the commonly used pre-processing~\cite{bib:NIPSWorkshop17:Paszke}, each sample (single image) is randomly resized and cropped into a $224\times224$ color image.
We use the original training dataset for training, and randomly select 10000 samples in the original validation dataset for validation, and the rest 40000 samples for testing.
We use a mini-batch with a size of 1024 training on 4 NVIDIA V100 GPUs. 

\fakeparagraph{ResNet50.}
We use pytorch-style ResNet50, which is slightly different than the original Resnet-50~\cite{bib:CVPR16:He}.
The down-sampling (stride=2) is conducted in $3\times3$ conv layer instead of $1\times1$ conv layer.
The network architecture is the same as ``resnet50'' in~\cite{bib:torchResNet}.
ResNet50 needs around 102.23MB to store the weights and around 4089MFLOPs for a single image inference.
We use the Nesterov SGD optimizer with the cosine schedule for learning rate decay. 
The initial learning rate is set as 0.5; the momentum is set as 0.9; the weight decay is set as 0.0001; the number of training epochs is set as 150.

\section{Additional Experiments}
\label{app:experiments}

\subsection{Iterative Training vs. Parallel Training}
\label{app:iter_para}

In this part, we compare \sysname (parallel training) with iterative training multiple subnets.
We implement two iterative training methods mentioned in \secref{sec:optimize}, namely iterative training with progressively increased/decreased sparsity (see Alg.~\ref{alg:increased} and Alg.~\ref{alg:decreased} in Appendix~\ref{app:iterative} respectively).
The loss of each subnet is optimized separately in iterative training. 
Thus for a fair comparison, we do not re-weight loss in \sysname, \ie $\gamma=0$.
Also in all experiments, we conduct unstructured sampling in the entire tensor, and allow BN layers to be fine-tuned individually for each subnet to avoid other side effects.

The comparison results are plotted in \figref{fig:parallel} Left. 
Parallel training substantially outperforms iterative training. 
Iterating over increased sparsity does not provide any space to optimize subnets with higher sparsity. 
Therefore, the accuracy drops quickly along iterations.
Although iterating over decreased sparsity may yield a well-performed high sparsity network, the accuracy does not improve significantly afterwards. 
We argue this is due to the fact that iterative training causes the optimizer to end in a hard to escape region around the previous subnet in the loss landscape. 
On the contrary, parallel training allows multiple subnets to be sampled and optimized jointly, which may especially benefit highly sparse networks, see \figref{fig:parallel} Left. 

\subsection{Correction factor $\gamma$}
\label{app:gamma}

The loss weights $\pi_k$ used in the parallel training may influence the final accuracy of different subnets.
In \secref{sec:optimize}, we introduce a correction factor $\gamma$ to control $\pi_k$. 
We thus conduct a set of experiments with different $\gamma$.
$\gamma=0$ means all loss items are weighted equally; $\gamma>0$ means the loss of the lower sparsity subnets is weighted larger, and vice versa. 
For example, for ResNet20 with $s_{1:5}=0.8,0.9,0.95,0.98,0.99$, when $\gamma=0.5$, $\pi_{1:5}\approx0.36,0.26,0.18,0.12,0.08$; $\gamma=-1.0$, $\pi_{1:5}\approx0.03,0.05,0.11,0.27,0.54$.

\begin{figure}[tbp!]
    \vspace{-1em}
	\centering
	\includegraphics[width=0.47\textwidth,height=0.20\textwidth]{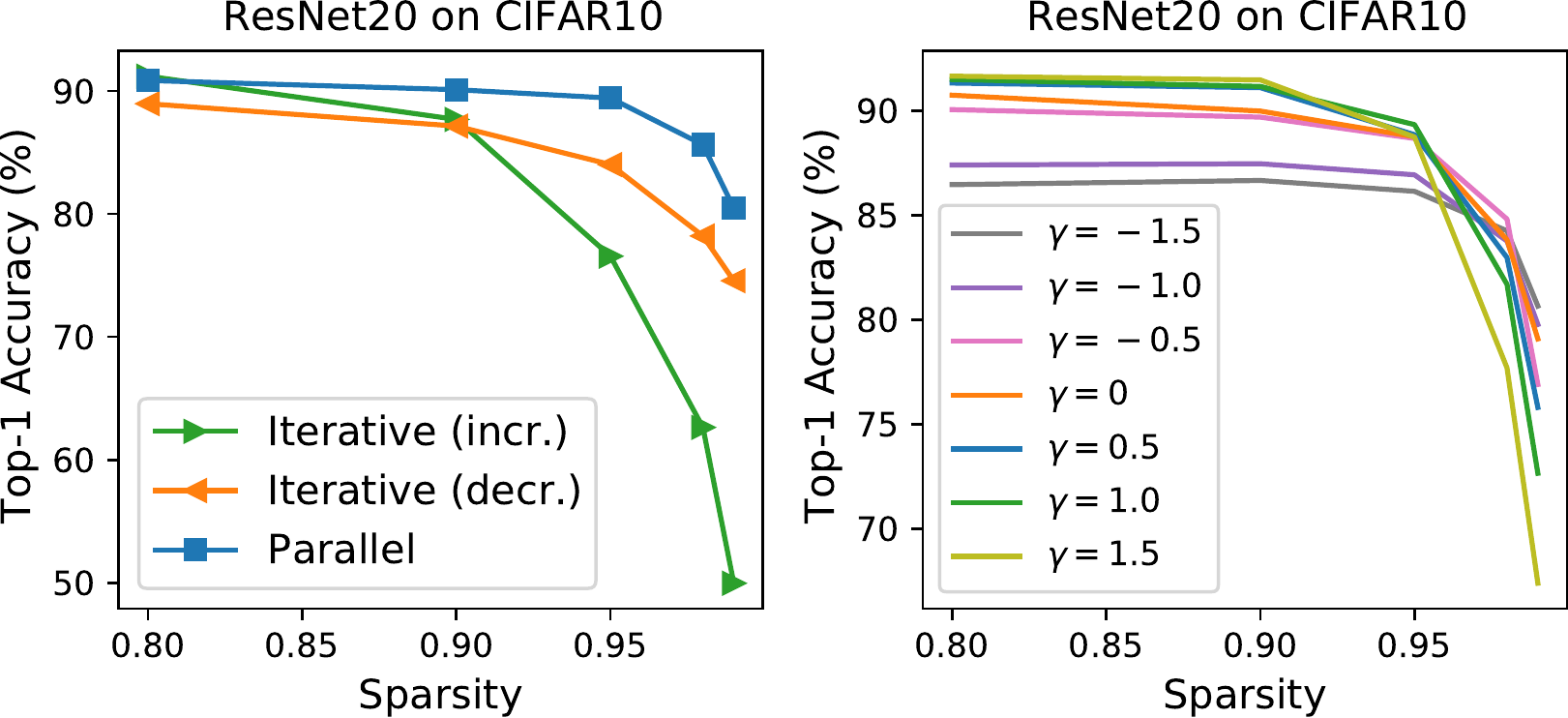}
	\caption{Left: Comparing parallel training with iterative training. Right: Ablation studies on the correction factor $\gamma$.}
    \label{fig:parallel}
	\vspace{-1.2em}
\end{figure}

The results in \figref{fig:parallel} Right show that the high sparsity subnets generally yield a higher final accuracy with a smaller $\gamma$. 
This is intuitive since a smaller $\gamma$ assigns a larger weight on the high sparsity subnets.
However, the downside is that the most powerful subnet (with the lowest sparsity) can not reach its top accuracy.
Note that the most powerful subnet is often adopted either under the critical case requiring high accuracy or in the commonly used scenario with standard resource constraints, see in \secref{sec:introduction}.
Also as discussed in \secref{sec:optimize}, low sparsity subnets should be weighted more, since they are implicitly optimized with a smaller step size.
Experimentally, we find that $\gamma\in[0.5,1]$ in parallel training allows us to train a group of subnets where the most powerful subnet can reach a similar accuracy as training in separately. 
We set $\gamma=0.5$ in the experiments.

\subsection{Cosine Similarity Analysis}
\label{app:cosine}

\begin{figure*}[htbp!]
	\centering
	\includegraphics[width=0.6\textwidth]{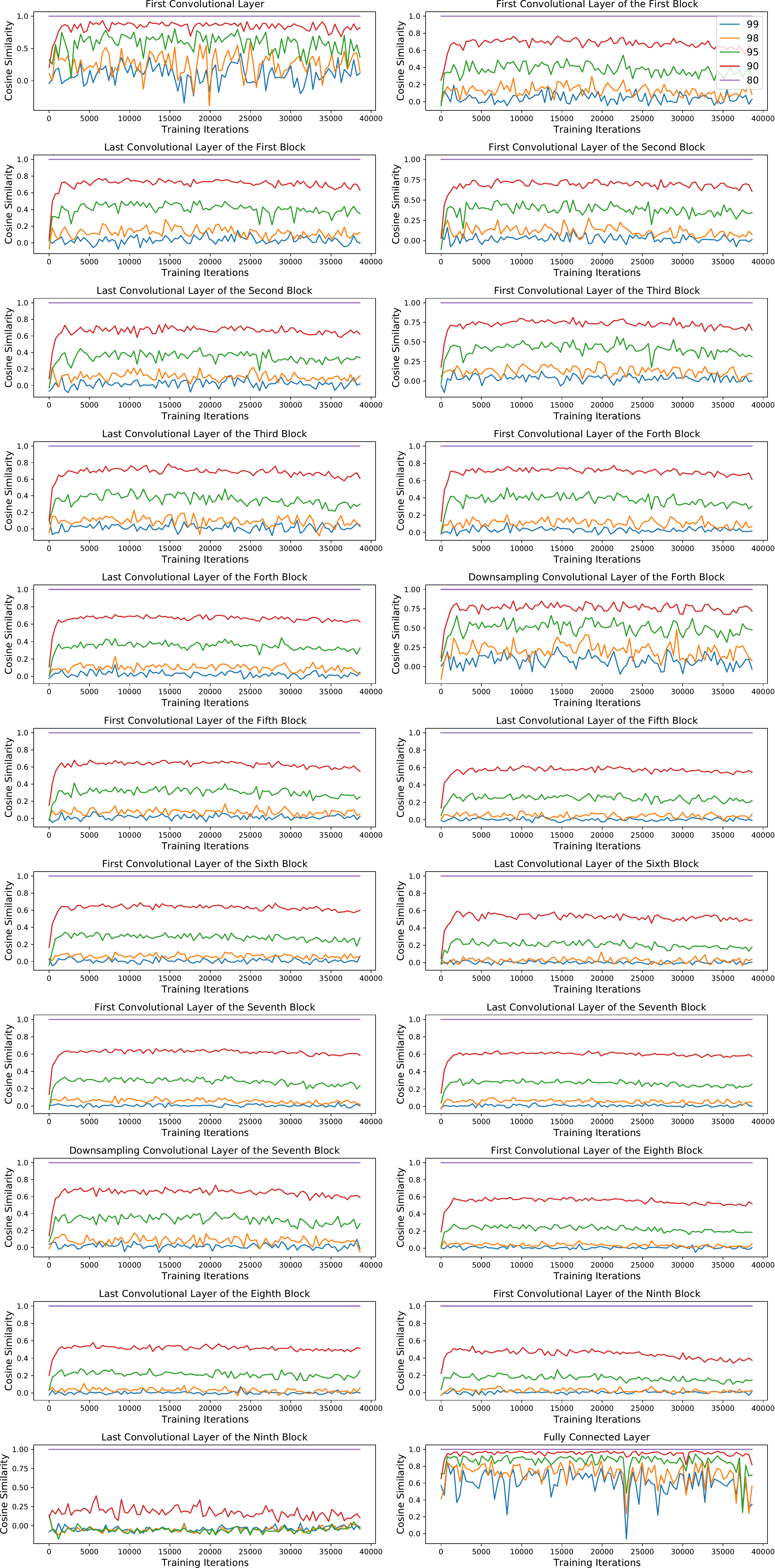}
	\caption{The cosine similarity between the loss gradients of 5 subnets (with sparsity $0.8,0.9,0.95,0.98,0.99$) and that of the lowest sparsity subnet (with sparsity 0.8) along with the parallel training iterations. We show all conv layers and the fc layer in ResNet20.}
	\label{fig:cosine_appendix}
\end{figure*}

In \figref{fig:process} Right of \secref{sec:optimize} in the main text, we choose two typical layers in ResNet20 and plot their cosine similarity between the loss gradients of different subnets along with training iterations.
Here, we show the cosine similarity between the loss gradients of all layers in ResNet20, see in \figref{fig:cosine_appendix}.
Recall that we parallelly train 5 subnets of ResNet20 \cite{bib:CVPR16:He} with the overall sparsity $s_{1:5}=0.8,0.9,0.95,0.98,0.99$ on CIFAR10, and let the 5 loss items weighted equally, \ie $\pi_{1:5}=0.2$. 
We plot the cosine similarity between the loss gradients of 5 subnets (\ie $(\partial\mathcal{L}(\bm{w}_k)/\partial\bm{w}_k) \odot \bm{m}_k$ with $k=1,...,5$) and that of the lowest sparsity subnet (\ie $(\partial\mathcal{L}(\bm{w}_1)/\partial\bm{w}_1) \odot \bm{m}_1$) along with the training iterations.
It shows that the loss gradients of different subnets are always positively correlated with each other.
The results also verify that multiple subnets are jointly trained towards the optimal point in the loss landscape.

\end{document}